\documentclass{article}
\usepackage{ijcai23}

\usepackage{times}
\usepackage{soul}
\usepackage{url}
\usepackage[title]{appendix}
\usepackage[hidelinks]{hyperref}
\usepackage[utf8]{inputenc}
\usepackage[small]{caption}
\usepackage{graphicx}
\usepackage{amsmath}
\usepackage{amsthm}
\usepackage{booktabs}
\usepackage{algorithm}
\usepackage{algorithmic}
\usepackage[switch]{lineno}
\usepackage{todonotes}
\usepackage{multirow}
\usepackage{xspace}
\usepackage{arydshln}
\usepackage{subcaption}

\usepackage{bm}

\def\eqref#1{equation~\ref{#1}}

\def\1{\bm{1}}

\DeclareMathAlphabet{\mathsfit}{\encodingdefault}{\sfdefault}{m}{sl}
\SetMathAlphabet{\mathsfit}{bold}{\encodingdefault}{\sfdefault}{bx}{n}

\usepackage{array}

\definecolor{maroon}{RGB}{191, 96, 96}

\definecolor{purple}{RGB}{100, 0, 200}

\definecolor{mint}{rgb}{0.24, 0.71, 0.54}

\definecolor{mint}{rgb}{0.24, 0.71, 0.54}

\newcolumntype{L}[1]{>{\raggedright\let\newline\\\arraybackslash\hspace{0pt}}m{#1}}
\newcolumntype{C}[1]{>{\centering\let\newline\\\arraybackslash\hspace{0pt}}m{#1}}
\newcolumntype{R}[1]{>{\raggedleft\let\newline\\\arraybackslash\hspace{0pt}}m{#1}}

\makeatletter
\def\adl@drawiv#1#2#3{%
        \hskip.5\tabcolsep
        \xleaders#3{#2.5\@tempdimb #1{1}#2.5\@tempdimb}%
                #2\z@ plus1fil minus1fil\relax
        \hskip.5\tabcolsep}
\newcommand{\cdashlinelr}[1]{%
  \noalign{\vskip\aboverulesep
           \global\let\@dashdrawstore\adl@draw
           \global\let\adl@draw\adl@drawiv}
  \cdashline{#1}
  \noalign{\global\let\adl@draw\@dashdrawstore
           \vskip\belowrulesep}}
\makeatother

\newcommand{\cmmnt}[1]{\ignorespaces}

\newcommand{\modelbaseline}{\textsc{Base}\xspace}
\newcommand{\modelbaselinedebias}{\textsc{Base-CDA}\xspace}
\newcommand{\modelours}{\textsc{Ours}\xspace}
\newcommand{\modeloursdebias}{\textsc{Ours-CDA}\xspace}

\newcommand{\modelxlm}{XLM-R\xspace}
\newcommand{\modelmbert}{m-BERT\xspace}
\newcommand{\modelhumbert}{HumBERT\xspace}

\newcommand{\humset}{\textsc{HumSet}\xspace}
\newcommand{\humsetbias}{\textsc{HumSetBias}\xspace}

\title{Leveraging Domain Knowledge for Inclusive and Bias-aware Humanitarian Response Entry Classification}

\author{
Nicolò Tamagnone$^{1,2}$  \and
Selim Fekih$^1$ \and
Ximena Contla$^1$  \and
Nayid Orozco$^1$  \and
Navid Rekabsaz$^3$
\affiliations
$^1$Data Friendly Space\\
$^2$ISI Foundation, Turin, Italy\\
$^3$Johannes Kepler University Linz, LIT AI Lab, Austria
\emails
$^1$\{nico,selim,ximena,nayid\}@datafriendlyspace.org,
$^3$navid.rekabsaz@jku.at
}

\begin{document}

\maketitle

\begin{abstract}
Accurate and rapid situation analysis during humanitarian crises is critical to delivering humanitarian aid efficiently and is fundamental to humanitarian imperatives and the Leave No One Behind (LNOB) principle. This data analysis can highly benefit from language processing systems, e.g., by classifying the text data according to a humanitarian ontology. However, approaching this by simply fine-tuning a generic large language model (LLM) involves considerable practical and ethical issues, particularly the lack of effectiveness on data-sparse and complex subdomains, and the encoding of societal biases and unwanted associations. In this work, we aim to provide an effective and ethically-aware system for humanitarian data analysis. We approach this by (1) introducing a novel architecture adjusted to the humanitarian analysis framework, (2) creating and releasing a novel humanitarian-specific LLM called \modelhumbert, and (3) proposing a systematic way to measure and mitigate biases. Our experiments' results show the better performance of our approach on zero-shot and full-training settings in comparison with strong baseline models, while also revealing the existence of biases in the resulting LLMs. Utilizing a targeted counterfactual data augmentation approach, we significantly reduce these biases without compromising performance. 

\end{abstract}

\section{Introduction}
\label{sec:introduction}

The efficiency of humanitarian aid during crises, be it natural disasters, conflicts, or pandemics such as COVID-19, relies heavily on quick and accurate analysis of relevant data within the first 72 hours of a disaster. During these crucial hours, humanitarian response analysts from international organizations and non-governmental organizations (NGOs) sift through a large volume of data, related to the crisis, to obtain a comprehensive understanding of the situation. A significant amount of this information is from secondary sources such as reports, news articles, and text-based data. During this phase, the process of analysis involves extracting critical information from those sources and categorizing it based on established humanitarian frameworks and guidelines. The analysis of this information plays a crucial role in determining the appropriate relief efforts to implement, and can highly benefit from technological solutions based on natural language processing (NLP) and Deep Learning. In line with the Leave No One Behind (LNOB) principle -- the central and transformative promise of the 2030 Agenda for Sustainable Development Goals -- such systems can assist humanitarian organizations with efficient and accurate information analysis while supporting the generation of evidence and data disaggregation that goes beyond gender, geography, and age.

To enable training and benchmarking NLP models in the humanitarian domain, the recently released \humset dataset~\cite{fekih-etal-2022-humset} provides a rich resource of humanitarian document analysis and entry (text excerpt) classification. The dataset originated from the Data Entry and Exploration Platform (the DEEP)\footnote{\url{https://thedeep.io/}} developed and maintained by Data Friendly Space, a U.S.-based international non-governmental organization (INGO),\footnote{\url{https://datafriendlyspace.org/}} in collaboration with several international organizations.\footnote{Such as the International Federation of Red Cross (IFRC), the United Nations High Commissioner for Refugees (UNHCR), and the United Nations Office for the Coordination of Humanitarian Affairs (UNOCHA), DEEP is funded by USAID's Bureau for Humanitarian Assistance (BHA).} \humset is a multi-lingual dataset, covering several disasters around the globe, where the informative entries of each document are extracted by humanitarian analysts, and classified/tagged according to a designated humanitarian information analysis framework. Fekih et al.~\shortcite{fekih-etal-2022-humset} further use this data to fine-tune large pre-trained language models (LLMs) such as {XLM-RoBERTa}~\cite{xlmr}, observing the overall effectiveness of these baseline approaches. However, we spot several challenges and concerns regarding this approach, especially when one aims to utilize these models in the real workflow of humanitarian response. These challenges are due to the inherent complexity of the humanitarian framework, the lack of training data on new crises, and the presence of societal biases and unwanted associations in the models. Our aim in this work is to address these challenges and provide an effective and ethically-aware solution for humanitarian response entry classification, as explained in the following.  

The first challenge is rooted in the fact that the various categories of the humanitarian framework share topical conceptual similarities, while each category also requires its own learning capacities. Learning all categories together in one model -- as done in the baseline approach -- leads to overwriting of information and catastrophic forgetting, as similarly reported in the context of multi-task learning~\cite{parisi2019continual,chen-etal-2020-recall}. We approach this challenge by taking into account the composition of the humanitarian framework and introduce a combinatorial model consisting of several shared layers across categories/tasks, as well as task-specific Transformer~\cite{NIPS2017_3f5ee243} layers. Our experiments, using various backbone LLMs, show the overall better performance of our proposed architecture in comparison with the baseline approaches.

The second challenge regards the (expected) weak performance of the models in the cases with no or little training data available (zero-/few-shot setting). This scenario is particularly common in the humanitarian domain, as each crisis -- while commonly happening abruptly and unexpectedly -- often differs from the previous incidents. We approach this challenge by creating and releasing a novel LLM called \modelhumbert, tailored to the characteristics of the humanitarian domain. Our experiments show that utilizing \modelhumbert significantly improves the performance on both zero-shot entry classification using prompting, and the standard fully-supervised setting when compared with the various baseline models using generic LLMs.

Finally, the third challenge concerns the issue of encoding societal biases and stereotypes in LLMs~\cite{sheng2019woman,blodgett2020language,rekabsaz2021measuring}, and carrying and exaggerating these biases in downstream tasks~\cite{rekabsaz2020neural,rekabsaz2021societal,kumar2023parameter,hauzenberger2023modular}. This issue is particularly critical, as a central principle in the humanitarian and development domains is to support minorities and underrepresented populations, in accordance with the LNOB principle, which might be neglected by a biased model. To address this challenge, we first create the \humsetbias dataset, a subset of \humset consisting of the data points that can potentially cause or reflect the biases with respect to gender and specific countries. Using \humsetbias, we then measure the gender/country bias in a model as the discrepancies of the model's prediction probabilities for a data point and its counterfactual form. Our results reveal the existence of a considerable amount of biases in the models and the effects of these biases on the decisions of the model. We further approach these biases using a counterfactual data augmentation (CDA)~\cite{zhao2018gender,lu2018gender} approach and achieve significant mitigation in the biases without any decrease in overall performance.  

The remainder of the work is organized as follows: in Section~\ref{sec:related}, we provide the background on the humanitarian domain and review the related work. Section~\ref{sec:model} explains the proposed architecture, followed by Section~\ref{sec:resources} to introduce the novel LLM and dataset resources. We explain the setup of our experiments and report their results in Section~\ref{sec:results}. Our code and models are available at \textbf{\url{https://github.com/the-deep-nlp/bias-aware-humanitarian-entry-classification}}.

\section{Background and Related Work}
\label{sec:related}

\begin{table}[t]
\centering
\scriptsize
\begin{tabular}{L{1.5cm} L{6.5cm}}
\toprule
\multicolumn{2}{l}{\textbf{Single-leveled Tasks:}} \\
\textbf{Sectors (11)} & Agriculture, Cross-sector, Education, Food Security, Health,  Livelihoods, Logistics, Nutrition, Protection, Shelter, WASH (Water, Sanitation \& Hygiene)\\ \cdashlinelr{1-2}  
\textbf{Pillars~1D (7)} & Context, COVID-19, Displacement, Humanitarian Access, Information \& Communication, Casualties, Shock/Event  \\ \cdashlinelr{1-2}  
\textbf{Pillars~2D (6)} & Capacities \& Response, Humanitarian Conditions, Impact, At Risk, Priority Needs, Priority Interventions  \\ \midrule

\multicolumn{2}{l}{\textbf{Two-leveled Tasks:}} \\
\multirow{7}{*}{\shortstack{\textbf{Subpillars~1D}\\\textbf{(36)}}} & \textit{Casualties:} Dead, Injured, Missing \\
& \textit{Context:} Demography, Economy, Environment, Legal \& Policy, Politics, Security \& stability, Socio-cultural  \\
& \textit{COVID-19:} Cases, Contact tracing, Deaths, Hospitalization \& care, Prevention campaign, Research and outlook, Restriction measures, Testing, Vaccination  \\
& \textit{Displacement:} Intentions, Local integration, Pull factors, Push factors, Type/numbers/movements  \\
& \textit{Humanitarian Access:} People facing humanitarian access constraints/humanitarian access gaps, Physical constraints, Population to relief, Relief to population  \\
& \textit{Information and Communication:} Communication means and preferences, Information challenges and barriers, Knowledge and info gaps (hum), Knowledge and info gaps (pop)  \\
& \textit{Shock / Event:} Hazard \& threats, Mitigating factors, Type and characteristics, Underlying\/aggravating factors  \\ \cdashlinelr{1-2}
 
\multirow{5}{*}{\shortstack{\textbf{Subpillars~2D}\\\textbf{(19)}}}  & \textit{At Risk:} Number of people at risk, Risk and vulnerabilities  \\ 
& \textit{Capacities \& Response:}  International response, Local response, National response, People reached/response gaps, Red cross/red crescent \\ 
& \textit{Humanitarian Conditions:} Coping mechanisms, Living standards, Number of people in need, Physical and mental well-being  \\
& \textit{Impact:} Driver/aggravating factors, Impact on people, Impact on systems, services, and networks, Number of people affected  \\ 
& \textit{Priority Needs:} Expressed by humanitarian staff, Expressed by population \\ 
& \textit{Priority Interventions:} Expressed by humanitarian staff, Expressed by population  \\ 
 \bottomrule
\end{tabular}
\caption{Overview of the humanitarian analysis framework. The number of tags present for each category is reported in parentheses.}
\vspace{-4mm}
\label{tbl:framework:overview}
\end{table}

\begin{figure*}[t]
    \centering
  \begin{subfigure}[b]{0.45\textwidth}
     \centering
     \includegraphics[width=\textwidth]{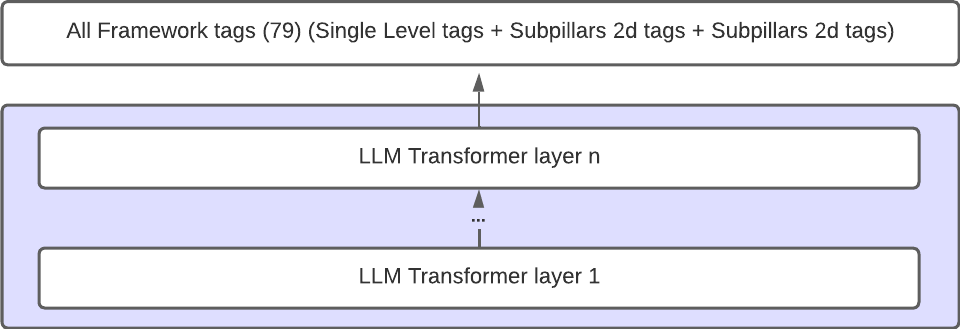}
     \caption{Baseline architecture}
     \label{fig:models:baseline}
  \end{subfigure}
  \hfill
  \begin{subfigure}[b]{0.45\textwidth}
     \centering
     \includegraphics[width=\textwidth]{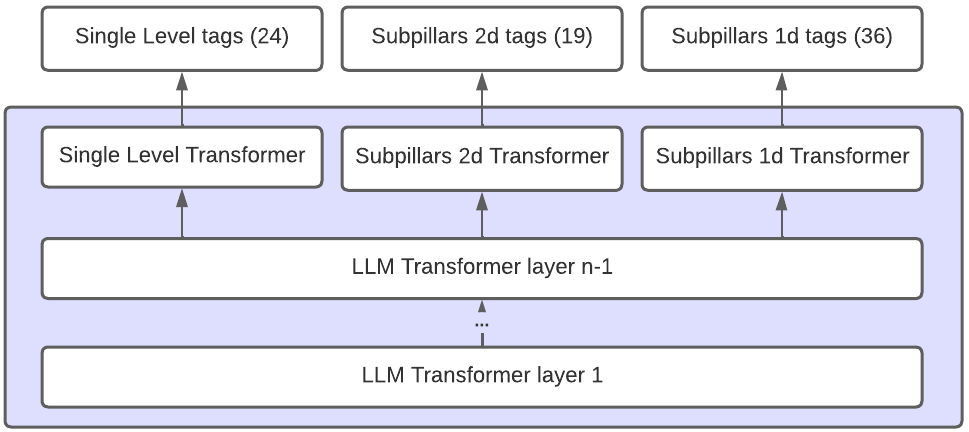}
     \caption{Our proposed architecture}
     \label{fig:models:ours}
  \end{subfigure}
  \caption{Model architectures used in this study}
  \label{fig:models}
  \vspace{-4mm}
\end{figure*}


Previous work introduces a number of resources for humanitarian data analysis. The \humset dataset~\cite{fekih-etal-2022-humset} -- used in the work at hand -- is created by humanitarian analysts on official documents and news from the most recognized humanitarian agencies. These data are organized according to the proposed analytical framework in the humanitarian domain, which defines a set of guidelines and structures to collect and organize data. This humanitarian analytical framework -- shown in Table~\ref{tbl:framework:overview} -- consists of five categories of Sectors, Pillars/Subpillars~1D, and Pillars/Subpillars~2D. The Subpillars~1D and Subpillars~2D categories are organized into a two-leveled tree hierarchy (Pillars to Subpillars). 

As other resources in this domain, Alam et al.\shortcite{alam2021crisisbench} provide CrisisBench, a benchmark of social media data labeled with the relevance of the text to humanitarian action, beneficial for filtering social media content as an additional resource of situational awareness. Similarly, Imran et al.~\shortcite{imran2016twitter} release human-annotated Twitter corpora accumulated during various crises in the span of 2013 to 2015, and \cite{9137480} and later \cite{alharbi-lee-2021-kawarith} publish Arabic Twitter classification datasets for crisis events. 

Using deep learning methods, Yela-Bello et al.~\shortcite{yela-bello-etal-2021-multihumes} study neural extractive summarization techniques to spot informative excerpts in humanitarian data. 
More recently, Lai et al.~\shortcite{lai2022natural} propose a hybrid named-entity recognition model that utilizes a set of features to extract the flooding information and risk reduction projects from newspapers. Our work contributes to this line of research by introducing a novel humanitarian-specific LLM leveraged in zero-shot and low-resource settings and approaching underlying biases.

\vspace{-2mm}
\section{Model Architecture}
\label{sec:model}

In this section, we explain the design of our model architecture. Our aim is to provide an effective (multi-label) entry classification model, which takes into account the complexity of the analytical framework consisting of categories with many labels (see Table \ref{tbl:framework:overview}), as well as the possible inter-category conceptual relationships. In the following, we first explain the baseline entry classification model and then extend it to our proposed approach. Both architectures leverage the power of pre-trained LLMs as the backbone. The schematic view of the two architectures is shown in Figure~\ref{fig:models}.

The baseline approach follows the standard classification approach of LLMs~\cite{devlin-etal-2019-bert}, namely  by first encoding the input sequence into a vector embedding, used for predicting the labels. Since the classification task is multi-label, a linear classification head is defined to predict each class (79 heads in total). The baseline architecture in fact does not take into account the hierarchy or the relationships of the framework's labels and treats all the categories with the same encoded embedding. 

Our proposed architecture aims to account for these aspects. Similar to the baseline, our approach shares $N-1$ Transformer layers of the LLM across all the categories, while the $N$th layer is replicated three times, forming three sub-layers. The first sub-layer is assigned to the single-leveled tasks namely Sectors, Pillars~1D, and Pillars~2D, and each of the other two sub-layers is dedicated to each of the two-leveled tasks, Subpillars~1D and Subpillars~2D. A linear classification head is then defined on top of each of the Transformer sub-layers, predicting only the labels belonging to the corresponding task. The resulting architecture is in fact a combination of the shared parameters and specific components and follows the relations and hierarchy of the analytical framework's label space.

\vspace{-2mm}
\section{Novel Humanitarian Resources}
\label{sec:resources}
In this section, we present a suite of innovative resources designed to improve humanitarian response classification.  Our resources include \modelhumbert, a new LLM model that incorporates insights of the latest research on language modeling specifically trained on a vast corpus of humanitarian text data, and \humsetbias, a dataset for analyzing and mitigating biases in humanitarian response classification.

\subsection{\modelhumbert}
\label{sec:resource:humbert}
The advancement of LLMs in recent years have provided immense benefits to the field of NLP, particularly by enabling effective transfer learning approaches~\cite{malte2019evolution}. LLMs are trained on vast amounts of general text data, and have shown remarkable performance across a wide range of language downstream tasks. 
However, 
such general LLMs may not perform as well as on specific domains, containing particular topics, specialized vocabulary, and complex concepts not well-represented in general language texts. To address this challenge, recent works focus on developing pre-trained language models for different domains, as for instance SciBERT~\cite{scibert}, BioBERT~\cite{biobert}, and FinBERT~\cite{finbert} are trained on scientific, biomedical and financial textual data, respectively. In the work in hand, we introduce \modelhumbert, a multilingual large language model trained on a large corpus of humanitarian text data, including reports, news articles, and other materials related to humanitarian crises and aid efforts. 

\paragraph{Humanitarian Corpus Data Collection} The text data for creating the humanitarian corpus is collected from three main sources: ReliefWeb,\footnote{\url{https://reliefweb.int/}} UNHCR Refworld,\footnote{\url{https://www.refworld.org/}} and Europe Media Monitor News Brief (EMM).\footnote{\url{https://emm.newsbrief.eu/}} ReliefWeb is a humanitarian information portal that provides news, reports, maps, and data related to crises. UNHCR Refworld is a database mainly focused on refugee-related documentation, including country of origin information, policy documents, and legal analysis. EMM is a real-time news monitoring tool that aggregates articles and media posts, with a specific focus on issues related to the European Union. While the main focus of Reliefweb and UNHCR Refworld is related to humanitarian crises and refugees, EMM broadcasts a wider range of news articles. We hence apply a filtering criterion on themes such as \emph{Crisis Response}, \emph{Humanitarian Aid}, \emph{Food Security \& Food Assistance} and \emph{Asylum}, to retrieve the documents relevant to humanitarian domain.

We collect the mentioned text articles, resulting in a  collection of $\sim\!\!2$M documents, of which $90\%$ are published articles from 2003 to 2022. The documents are in three languages of English ($82\%$), French ($10\%$), and Spanish ($8\%$). We extract the text from the documents (web pages, related PDF attachments, etc.). The documents have on average $26$ sentences. 
We keep the structure of the text consistent with one of the paragraphs in the original document. We execute text cleaning and simple pre-processing of the text (details available in the code). After concatenating all text together, the resulting corpus consists of $\sim\!\!50$M sentences with a total size of approximately 2B tokens. We further process the corpus to remove personal information such as e-mails and phone numbers. 

\paragraph{Training} \modelhumbert is a domain-specific LLM that has been fine-tuned on our humanitarian corpus, starting from the checkpoint of {XLM-RoBERTa$_\mathrm{Base}$}~\cite{xlmr}. 
Starting from this model provides us the advantage of its pre-trained multilingual capabilities, which is then further fine-tuned to capture the linguistic nuances specific to the humanitarian domain. We train \modelhumbert for 3 epochs with $10\%$ of the corpus used as the validation set. Following the training procedure of XLM-RoBERTa$_\mathrm{Base}$,\footnote{https://huggingface.co/xlm-roberta-base} we further fine-tune the model based on its masked language modeling objective, using the training pipeline provided by the HuggingFace library~\cite{wolf-etal-2020-transformers}.\footnote{\url{https://github.com/huggingface/transformers/blob/main/examples/pytorch/language-modeling/run_mlm.py}} The training is performed on a single v2-8 TPU.\footnote{Research supported with Cloud TPUs from Google's TPU Research Cloud (TRC)}

\begin{table}[t]
\centering
\small
\begin{tabular}{lll}
\toprule
 Female &  Male &  Neutral \\
\midrule
 she     &     he   &  they \\
 woman   &    man   &  person, individual \\
 women   &    men   &  persons, individuals \\
 mother  & father   &  person, individual \\
 mothers & fathers  &  persons, individuals \\ 
 girl    &    boy   &  child \\
 girls   &   boys   &  children \\
 her     &    his   &  their \\
 her     &    him   &  them \\
 female  &   male   & person, individual \\
 females &   males  &  persons, individuals \\
 wife    &  husband &  person, individual \\
 wives   & husbands &  persons, individuals \\ 
\bottomrule
\end{tabular}

\caption{Keywords indicating gender information, and the corresponding mappings for creating counterfactual samples.}
\label{table:gender-mappings}
\vspace{-4mm}
\end{table}

\vspace{-1mm}
\subsection{\humsetbias}
\label{sec:resource:humsetbias}
\vspace{-1mm}

As described in Fekih et al.~\shortcite{fekih-etal-2022-humset} and also touched upon in the previous sections, \humset dataset is created by the teams of humanitarian experts in various international organizations, through annotating the relevant entries of a text document according to the analysis framework. However, the distribution of the annotations across the labels/categories is influenced by the types of resources and the information involved. Given the diverse range of humanitarian crises and topics covered by \humset, the annotations might be skewed toward the context in which the contents are produced. As a result, the annotated data -- and hence the models trained on the dataset -- may exhibit biases towards attributes such as gender and country (geographical), which could impact the accuracy and fairness of downstream applications. 

To enable measuring and (potentially) addressing these biases, we introduce \humsetbias, a subset of the English part of the \humset dataset, created by searching for specific sensitive English keywords related to genders and countries within the annotated text. In addition, we extended this subset by incorporating \emph{targeted} counterfactual~\cite{roese1995counterfactual} samples, generated by modifying the original entries in order to create the altered versions of each text with gender/country information. The purpose of \humsetbias is to provide a more targeted resource for analyzing and addressing potential biases in humanitarian data and to enable the development of accurate and bias-aware NLP applications in the humanitarian sector. In the following, we explain the process of creating \humsetbias, how it is extended with targeted counterfactual samples.

\begin{table}[t]
\centering
\small

\begin{tabular}{l l rrrr}
\toprule
Bias Attribute & Bias label & Train & Val. & Test & All \\\midrule
\multirow{4}{*}{Gender} & Female  & 1,173 & 92 & 80 & 1,345\\
                          & Male    & 604 & 57 & 65 & 726\\
                          & Neutral & 15,271 & 1,443 & 1,524 & 18,238\\\cdashlinelr{2-6}
                          & Sum     & 17,048 & 1,592 & 1,669 & 20,309\\\midrule

\multirow{3}{*}{Country} & Venezuela & 261 & 20 & 13 & 294\\
                          & Syria     & 1,686 & 182 & 187 & 2,055\\
                          \cdashlinelr{2-6}
                          & Sum       & 1,947 & 202 & 200 & 2,349\\
 \bottomrule
\end{tabular}
\caption{Statistics of the \humsetbias dataset for each bias type, as subsets of the train/validation/test sets of \humset.}
\label{table:count-biases}
\vspace{-4mm}
\end{table}

\begin{table*}[t]
\centering
\begin{tabular}{l r r r r r r r r r r | r r}
\toprule
\multirow{2}{*}{LLM} & \multicolumn{2}{c}{Sectors} & \multicolumn{2}{c}{Pillars~1D} & \multicolumn{2}{c}{Subpillars~1D} & \multicolumn{2}{c}{Pillars~2D} & \multicolumn{2}{c}{Subpillars~2D}  & \multicolumn{2}{c}{Avg.} \\
 & Prec. & F1 & Prec. & F1 & Prec. & F1 & Prec. & F1 & Prec. & F1 & Prec. & F1 \\ \midrule
\modelmbert & 0.16 & 0.10 & 0.18 & 0.16 & 0.07 & 0.05 & 0.25 & \textbf{0.21} & 0.08 & 0.05 & 0.15 & 0.11\\
\modelxlm  & 0.27 & 0.23 & 0.24  & 0.21  & 0.11 & 0.09 & \textbf{0.26} &  0.18 & 0.09  & 0.06  & 0.19 & 0.15 \\
\modelhumbert & \textbf{0.28} & \textbf{0.24} & \textbf{0.27}  & \textbf{0.24} & \textbf{0.13}  &  \textbf{0.10} & 0.22  & 0.17  & \textbf{0.10}  &  \textbf{0.08}  & \textbf{0.20} & \textbf{0.17} \\
 \bottomrule
\end{tabular}
\vspace{-2mm}
\caption{Evaluation results of zero-shot classification on the five categories/tasks of the \humset dataset according to Precision and F1 metrics. ``Avg.'' reports the (macro) average of the results over the tasks. The best results in each task-metric are shown in bold.}
\label{table:zeroshot}
\vspace{-1mm}
\end{table*}

\begin{table*}[t]
\centering
\begin{tabular}{ll cc cc cc cc cc | cc}
\toprule
\multirow{2}{*}{LLM} & \multirow{2}{*}{Model} & \multicolumn{2}{c}{Sectors} & \multicolumn{2}{c}{Pillars 1D} & \multicolumn{2}{c}{Subpillars 1D} & \multicolumn{2}{c}{Pillars 2D} & \multicolumn{2}{c}{Subpillars 2D} & \multicolumn{2}{c}{Avg.}\\ 
& & Prec. & F1 & Prec. & F1 & Prec. & F1 & Prec. & F1 & Prec. & F1 & Prec. & F1\\ \midrule

\multirow{2}{*}{\modelmbert} & 
\modelbaseline & 0.75 & 0.77 & 0.68 & \textbf{0.70} & 0.51 & 0.51 & 0.7 & 0.69 & 0.54 & 0.48 & 0.71 & 0.72 \\ 
& \modelours   & 0.76 & 0.76 & 0.65 & 0.68 & 0.49 & 0.51 & 0.69 & 0.69 & 0.48 & 0.47 & 0.7 & 0.71 \\ \cdashlinelr{1-14} 

\multirow{2}{*}{\modelxlm} & 
\modelbaseline & 0.74 & 0.76 & \textbf{0.70} & 0.68 & 0.48 & 0.49 & 0.69 & 0.68 & 0.51 & 0.47 & 0.71 & 0.71 \\ 
& \modelours & 0.75 & 0.77 & 0.68 & 0.69 & 0.5 & 0.51 & 0.69 & 0.7 & 0.51 & 0.48 & 0.71 & 0.72 \\ \cdashlinelr{1-14}

\multirow{2}{*}{\modelhumbert} &  
\modelbaseline & 0.76 & \textbf{0.78} & 0.69 & \textbf{0.70} & 0.52 & 0.53 & 0.7 & 0.7 & 0.52 & 0.5 & \textbf{0.72} & \textbf{0.73} \\ 
& \modelours   & \textbf{0.77} & \textbf{0.78} & 0.69 & \textbf{0.70} & \textbf{0.57} & \textbf{0.54} & \textbf{0.71} & \textbf{0.71} & \textbf{0.55} & \textbf{0.51} & \textbf{0.72} & \textbf{0.73} \\

 \bottomrule
\end{tabular}
\caption{Evaluation results on fully training the models on the \humset dataset. The best results across LLMs are shown in bold. }
\label{table:subpillars1d-nb}

\vspace{-4mm}
\end{table*}

\paragraph{Dataset Creation}
To study country bias, we select Syria and Venezuela, given their frequent occurrence in \humset and also the fact that these countries encompass distinct types of humanitarian crises, namely the Syrian civil war\footnote{\url{https://en.wikipedia.org/wiki/Syrian_civil_war}} and the Venezuelan socioeconomic and political crisis\footnote{\url{https://en.wikipedia.org/wiki/Crisis_in_Venezuela}}. Through keyword search, we selected the data points that contain the name of these countries. We further aim to ensure that the selected entries are focused on countries, and do not contain other information, for instance about the nationalities, population, or geographical locations inside the countries. To this end, we exclude entries that mention any nationality (including Venezuelan and Syrian), other countries, and all cities of the chosen countries. To identify the entries with potential biases towards genders, we create a list of keywords indicating female, male, and neutral terms, reported in Table~\ref{table:gender-mappings}. We use these keywords to select the gender-related subset of entries. After reviewing this subset, we exclude the entries that contain concepts specific to one gender (such as the ones with words like \emph{pregnant} or \emph{lactating}), as well as the entries referring to more than one gender. Table~\ref{table:count-biases} reports the statistics of the resulting potentially biased excerpts according to the train, validation, and test sets, selected from the corresponding sets in \humset. 





\paragraph{Targeted Counter-factual Samples}
Various bias measurement and mitigation approaches can benefit from the existence of counterfactual samples, namely when the protected attribute of a given data point is represented in other possible forms. To enable this research, we additionally provide targeted counterfactual samples for each data point in \humsetbias dataset. In particular for gender attribute, we replace each female/male/neutral word with the two other cases (for instance $\text{female}\rightarrow\{\text{male},\text{neutral}\}$), according to the mapping in Table~\ref{table:gender-mappings}. We follow a similar procedure for the country-related data points by selecting Canada as a contrastive (neutral) case, as this country is not associated with any crisis in the \humset dataset. The counterfactual samples are hence created by replacing each source country in each data entry with the other source country and Canada, namely  $\text{Syria}\rightarrow\{\text{Venezuela},\text{Canada}\}$, and $\text{Venezuela}\rightarrow\{\text{Syria},\text{Canada}\}$. The data of targeted counterfactual samples is provided along with the \humsetbias dataset in our repository.

\vspace{-2mm}
\section{Results and Analysis}
\label{sec:results}

\begin{figure*}[t]
  \centering
  \begin{subfigure}[b]{1.0\textwidth}
    \centering
    \includegraphics[width=1\textwidth]{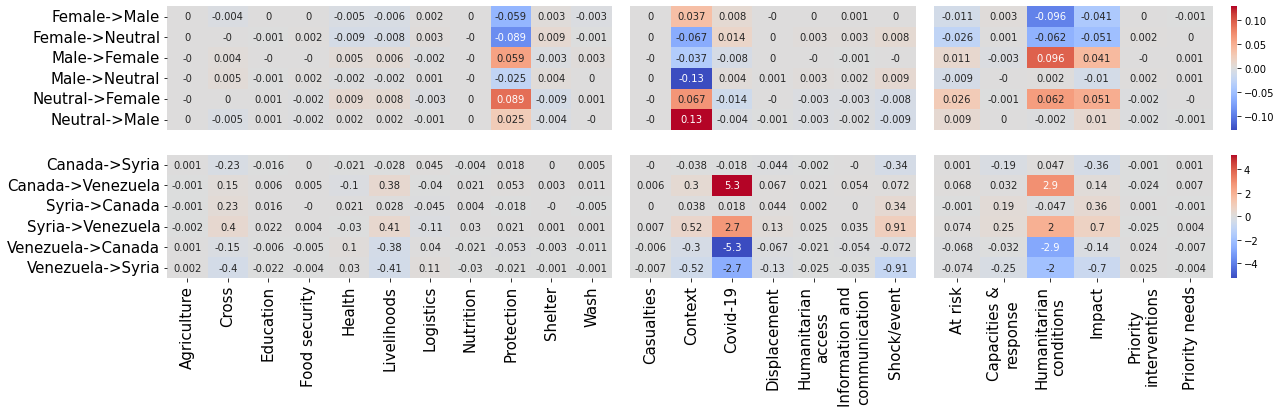}
    \label{fig:results:bias}
  \end{subfigure}
  
  \begin{subfigure}[b]{1.0\textwidth}
    \centering
    \includegraphics[width=1\textwidth]{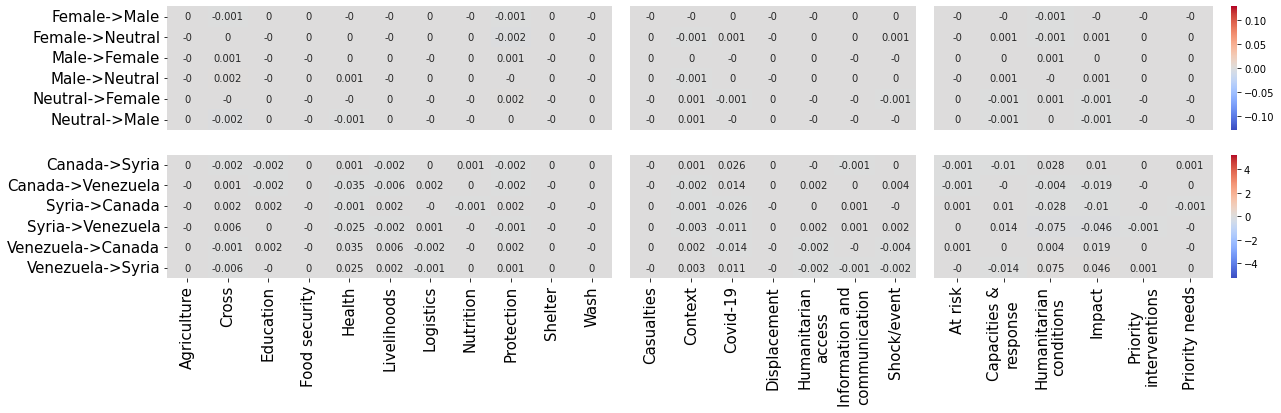}
    \label{fig:results:debias}
  \end{subfigure}
  \caption{
  The results of $\text{Tag-Shift}$ bias metric for \modelours architecture using the \modelhumbert as the backbone. (Top) Original model without debiasing; (Bottom) Counterfactual debiasing.
  }
  \label{fig:results-heatmap}
  \vspace{-4mm}
\end{figure*}

To conduct our experiments, we use the \humset dataset for training and evaluating the models on the classification task for the five humanitarian categories (see Section~\ref{sec:related}). 
We utilize three LLMs, namely {m-BERT}~\cite{devlin-etal-2019-bert}, {XLM-RoBERTa}~\cite{xlmr}, and HumBERT. Each LLM is used as the backbone for both the baseline architecture and our proposed one (see Section~\ref{sec:model}), referred to as \modelbaseline and \modelours, respectively. We fine-tune the hyperparameters on the validation set and report the evaluation results on the test set. Details of the models' hyperparameters are reported in Appendix~A.

In the following, we first explain our zero-shot classification method and the achieved results. We then report the performance of the models on the full-training scenario. Finally, leveraging the \humsetbias dataset we describe our method to measure the existence of gender/country bias in the models and report the results before and after applying the CDA bias mitigation method.

\subsection{Zero-shot Learning Results}

We utilize prompting to evaluate the abilities of the LLMs (in comparison with \modelhumbert) on entry classification zero-shot scenario. Previous studies have utilized prompts to perform downstream tasks by directly exploiting LLM objectives, showing promising results on few-/zero-shot settings \cite{schick-schutze-2021-exploiting,gpt}. As a common prompting method, the input text data point followed by a template with a masked token is given to an LLM to retrieve the probabilities of possible tokens to fill the mask. In our experiment, we utilize the simple template of ``\emph{It is about} \texttt{[MASK]}'', appended at the end of each input entry.  To perform classification concerning the labels of the humanitarian tasks (see Table \ref{tbl:framework:overview}), we create a \emph{verbalizer} to map the predicted \texttt{[MASK]} token to the labels. Previous work proposes various ways to formulate a verbalizer, namely manually~\cite{schick-schutze-2021-exploiting}, manually with pre-existing knowledge bases extensions~\cite{hu-etal-2022-knowledgeable}, or through automatic searching methods~\cite{gao-etal-2021-making} applicable in presence of training and validation examples. In our experiments, we follow the first approach and build our verbalizer manually, by associating each label --- whose class name is already semantically informative -- with a set of words with similar meanings or semantic relation. For example, the Sector label \emph{Agriculture} is associated with the terms \emph{farming}, \emph{cultivation}, \emph{land}, and \emph{farm}, in addition to the word \emph{agriculture}. The process results in an average of 4.6 words per label. 

Table~\ref{table:zeroshot} reports the results of the zero-shot classification on the five categories/tasks of the humanitarian analytical framework. The zero-shot prompting inference is performed on the test set of \humset for the three LLMs (without any training and validation). Based on the results, we observe that overall our domain-specific \modelhumbert outperforms the other two LLMs except for Pillars~2D. Generally, the XLM-R model shows better results than m-BERT, while this model is further improved by being fine-tuned on the humanitarian text data in \modelhumbert. One special case is the overall weak performance of all models on the Subpillars 1D/2D categories. We assume that this is due to the complexity associated with the semantics of the labels of these tasks, and the limitation of the manual verbalizer to comprehensively address the diverse facets of the meaning associated with these labels. We consider further investigation in this area as a line of research in our future work. 




\subsection{Full-training Performance}

We continue with evaluating the LLMs and the underlying model architectures (discussed in Section~\ref{sec:model}) on the fully supervised classification setting. The results are reported in Table~\ref{table:subpillars1d-nb}, where the best results per LLM are shown in bold, and the overall best results are indicated with underlining. 

As noted, our proposed architecture overall shows higher performance, particularly in the F1-scores. 
It is important to note that the results particularly increase for the Subpillars~1D task (especially for \modelxlm and \modelhumbert backbones), which is the category with the largest number of classes (36 tags). This observation supports the benefits of using the combination of shared and task-specific Transformer layers in our proposed architecture, to reduce the sparsity of classification heads without losing the shared knowledge across the categories. Comparing across the LLMs, \modelhumbert achieves the best performance in all the categories. 

\begin{table}[t]
\centering
\scalebox{0.91}{
\begin{tabular}{ll cc cc}
\toprule
\multirow{2}{*}{LLM} & \multirow{2}{*}{Model} & \multicolumn{2}{c}{Tasks Avg.$\uparrow$} & \multicolumn{2}{c}{Bias$\downarrow$} \\
& & Prec. & F1 & Gender & Country  \\ \midrule

\multirow{4}{*}{\modelmbert} & \modelbaseline & 0.71 & 0.72 & 0.57 & 9.79 \\ 
                            & \modelours      & \textbf{0.72} & 0.72 & 0.48 & 6.37  \\
                       & \modelbaselinedebias & 0.7 & 0.71 & 0.02 & 0.17  \\
                           & \modeloursdebias & \textbf{0.72} & 0.72 & 0.02 & 0.12  \\\cdashlinelr{1-6}

\multirow{4}{*}{\modelxlm} & \modelbaseline & 0.71 & 0.71 & 0.67 & 8.69 \\ 
                            & \modelours    & 0.71 & 0.72 & 0.4 & 3.67  \\
                     & \modelbaselinedebias & 0.71 & 0.72 & 0.05 & 0.14  \\
                         & \modeloursdebias & \textbf{0.72} & 0.72 & \textbf{0.01} & \textbf{0.11} \\\cdashlinelr{1-6} 

\multirow{4}{*}{\modelhumbert} & \modelbaseline & \textbf{0.72} & \textbf{0.73} & 0.71 & 8.37 \\ 
                               &   \modelours   & 0.71 & \textbf{0.73} & 0.28 & 6.52  \\
                         & \modelbaselinedebias & \textbf{0.72} & \textbf{0.73} & 0.02 & \textbf{0.11}  \\
                             & \modeloursdebias & \textbf{0.72} & \textbf{0.73} & \textbf{0.01} & 0.13  \\

 \bottomrule
\end{tabular}
}
\caption{Results of $\text{Overall-Shift}$ bias metric, and average tasks performance (taken from Table~\ref{table:subpillars1d-nb}). The best results per LLM are shown in bold.}
\label{table:overall-results}
\end{table}

\subsection{Bias Measurement and Mitigation}
\label{sec:results:bias}
To measure the bias of the models, 
we use \humsetbias to calculate the sensitivity of the trained models to the changes in gender and countries keywords concerning the changes in the predicted probabilities of the labels. A non-biased model should not be sensitive to such changes so that the predicted probabilities should remain the same when the respective words of  gender/country are swapped. 

To this end, our bias measurement is conducted on an extended version of the test sets, where each test set also includes the counterfactual samples of each data point (described in Section \ref{sec:resource:humsetbias}). Using this data, we first define the $\text{P-Shift}$ metric as the shift in the predicted probability of the entry $x$ concerning the tag $t$ with the bias label $m$ when replacing the label with $n$, formulated below:
\begin{equation}
\text{P-Shift}(x, t, m\rightarrow n)=\left(P(x_n, t) - P(x_m, t)\right) \times 100  
\end{equation}
Here, $P(x_m,t)$ is the prediction probability of $x$ on the tag $t$ in the form of bias label $m$, and $P(x_n,t)$ is the probability on the same tag $t$ of a corresponding counterfactual sample of $x$ swapped to bias label $n$.
The values are multiplied by 100 for improving readability. 
The total predicted probability shift from the attribute $m$ to the attribute $n$ for a tag $t$ is therefore calculated as the median of the probability shifts: 
\begin{equation}
\text{Tag-Shift}(t, m\rightarrow n) = \mathop{Median}_{x\in X_m}\Bigl\{\text{P-Shift}(x, t, m\rightarrow n)\Bigl\}
\end{equation}
where $X_m$ refers to the set of data points with the bias label $m$. In our formulations, we choose the median (instead of the mean) to mitigate the effect of the outliers in $\text{P-Shift}$ in the final statistic. We should note that since the test sets are augmented with counterfactual forms concerning each bias label, $X_m$ has the same size across various bias labels $m$ (size of the full test dataset). Figure~\ref{fig:results-heatmap} top reports the results of $\text{Tag-Shift}$ for both bias attributes, calculated on our proposed architecture with \modelhumbert as the backbone LLM, for all $m\!\rightarrow\!n$ cases of the single-leveled tags. The results without debiasing (top) reveal the existence of both gender and country biases in the model, particularly to specific tags. For example, in the Covid-19 Pillars 1D, $\text{Venezuela} \rightarrow \text{Canada}$ and $\text{Venezuela}\rightarrow \text{Syria}$ are negative (-5.3 and -2.7), indicating that the model assigns higher probabilities on this tag for the data points related to Venezuela compared to Syria and Canada. 

\paragraph{Bias Mitigation with Counterfactual Data Augmentation (CDA)}
To conduct bias mitigation with CDA, we extend the training set of \humsetbias data with the respective counterfactual data points concerning both gender and country. This results in new training set with 166,990 data points, which extends the original 129,268 training instances of \humset. 
Figure~\ref{fig:results-heatmap} bottom shows the results of applying CDA. The results of other combinations are provided in Appendix~B. Looking at the results, we observe a significant and consistent decrease in the degree of both gender and country bias. These results show the effectiveness of balancing the training data concerning the protected attributes.

\paragraph{Overall Probability Shift}
The $\text{Tag-Shift}$ metric provides useful information regarding the bias of a specific tag according to a specific counterfactual transition. However, we are also interested in an overall metric, enabling straightforward comparison across different models. We hence define the aggregation metric by summing over all tags the average of $\text{Tag-Shift}$ absolute values across all $m\rightarrow n$ transitions. 
This metric is formulated as follows: 
\begin{equation}
\text{Overall-Shift}=\frac{1}{|C|}\sum_{m \rightarrow n\in C}{\sum_{t\in T}{ |\text{Tag-Shift}(t, m\rightarrow n)|} } 
\end{equation}
where $C$ indicates the set of all possible bias label transitions. For both gender and country attributes, $|C|$ is equal to 6. The results of the $\text{Overall-Shift}$ metrics for all backbone LLMs, architectures, and debiasing settings are reported in Table~\ref{table:overall-results}. For a better comparison, the table also reports the average task performance, taken from Table~\ref{table:subpillars1d-nb}.
These results similarly indicate that despite the existence of biases in all models, the degrees of biases significantly decrease after applying CDA using the provided data of \humsetbias. This bias mitigation does not cause any drop in the models' performance 
while observing slight improvements in some cases. Among various models, the \modelhumbert LLM together with our architecture demonstrates effective results in terms of both classification performance (the highest classification results on F1-measure) and bias measurement (lowest gender bias). These observations indicate the benefits of using our proposed model with CDA, as an effective and bias-aware classification solution for the humanitarian sector.



\section{Conclusion}
During humanitarian crises, processing and classification of the data greatly assist analysts and decision-makers in gaining a comprehensive understanding of the situation in line with humanitarian imperatives and the Leave No One Behind (LNOB) principle. In this work, we propose a set of resources and methods for building a practice-oriented, effective, and bias-aware solution for entry classification in the humanitarian domain. We first introduce a novel parameter-sharing model architecture, designed according to the structure of the humanitarian analysis framework, followed by creating and releasing \modelhumbert -- a novel LLM adapted to the topics and nuances of this domain. Our experiments on the \humset dataset in both zero-shot and full-training settings, show the overall performance improvements of the models when using \modelhumbert in comparison with generic LLMs with comparable size, and also the overall benefits of our proposed architecture in comparison with baseline fine-tuning, particularly on the tasks with many target classes. In addition, we introduce the \humsetbias dataset, consisting of data points with specific gender and country attributes. Utilizing the counterfactual variations of the data points in \humsetbias, we observe the significant effects of these attributes on the predictions of the models, which we successfully mitigate utilizing a targeted data augmentation method.

\newpage

\section{Ethical Statement}

This work aims to promote more equitable and inclusive humanitarian response practices, by providing a sector-specific model and a comprehensive dataset for analyzing and mitigating biases. However, we emphasize that the proposed data, model, and classification architecture have inherent limitations explained below, which should be considered thoroughly,  particularly when they are utilized in practice. 

\humsetbias is created using rule-based mappings. Hence, we further review the entries to exclude the ones with negative keywords for both gender and country attributes and conduct several sanity checks iterations. However, the dataset should not be assumed as completely error-free, as there might be some minor shifts in the results we obtained. We indeed aim to continue improving the dataset, particularly through the common effort of the humanitarian community. 

Next, while the CDA method significantly reduces bias, the models still contain bias. We should consider that such biases may still lead to some issues depending on the usage scenario. 

Finally, the \humsetbias dataset does not cover all possible societal biases in the humanitarian domain. We hence encourage more effort to enlarge this dataset to other critical attributes such as specific needs groups, affected groups, and age categories. We believe that these features are central to any humanitarian technological solution in support of the analysis, particularly given the humanitarian imperatives of neutrality and impartiality and the LNOB principle part of the 2030 agenda.

\section*{Acknowledgments}
 We are immensely thankful to the DEEP users, taggers, and project owners for their dedicated work in making this dataset possible. Furthermore, we acknowledge the DEEP board members, which guidance has been crucial in shaping the DEEP. We express our sincere appreciation to the entire DFS team, the ToggleCorp team, ISI Foundation, and the Johannes Kepler University Linz, for their continuous support in enabling the use of NLP in the humanitarian community. Their collective contributions have been instrumental in the success of this research. Furthermore, we also express our sincere appreciation to the USAID's Bureau for Humanitarian Assistance (BHA) for their trust in DFS and their generous grant support that allows us to work on these topics.\\
NT gratefully acknowledges the support from the Lagrange Project of the ISI Foundation
funded by CRT Foundation. NT thanks Ciro Cattuto, Kyriaki Kalimeri, Yelena Mejova, and Rossano Schifanella for their supervision.


\bibliographystyle{named}
\bibliography{references}

\section*{Appendix}
\appendix
\section{Hyperparameters}
Table~\ref{table:hyperparameters} reports the results of hyperparameter tunning.
\begin{table}[h]
\centering
\small
\begin{tabular}{ L{4cm} L{4cm}}
\toprule
Hyperparameters & Values \\ \midrule
Number of Epochs & 3\\
Initial Learning Rate & 1e-4 \\
Dropout Rate & 0.2 \\
Train Batch Size       & 8     \\ 
Validation Batch Size  & 16    \\ 
Optimizer       & Adam Weight (with the standard Pytorch (https://pytorch.org/) hyperparameters)  \\ 
Learning Rate Scheduler  & Pytorch StepLR (with decay=0.4, step size=1)     \\ 
LLM input text max length  & 200    \\ 
Freezed LLM layers & LLM Embedding and first LLM layer \\
Decision boundary threshold & Finetuned differently for each training setup and tag on the best F1 score validation set after training (from 20 values ranging from the minimum to the maximum probability predicted for each tag). \\
\bottomrule
\end{tabular}
\caption{Hyperparameters used for finetuning Classification models and for Generating final predictions}
\label{table:hyperparameters}
\end{table}

\section{Additional Results}
\begin{figure*}[h]
  \centering
  \begin{subfigure}[b]{1.0\textwidth}
    \centering
    \includegraphics[width=1\textwidth]{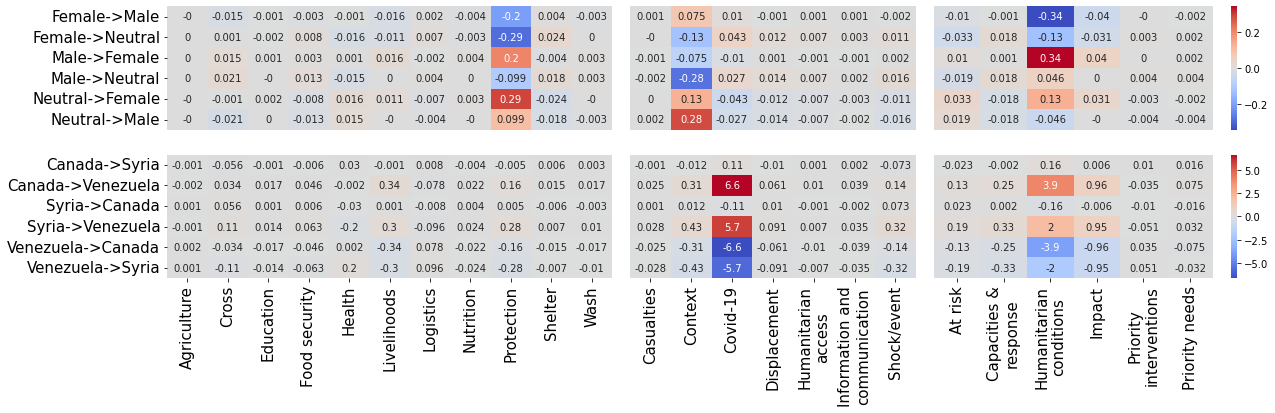}
  \end{subfigure}
  
  \begin{subfigure}[b]{1.0\textwidth}
    \centering
    \includegraphics[width=1\textwidth]{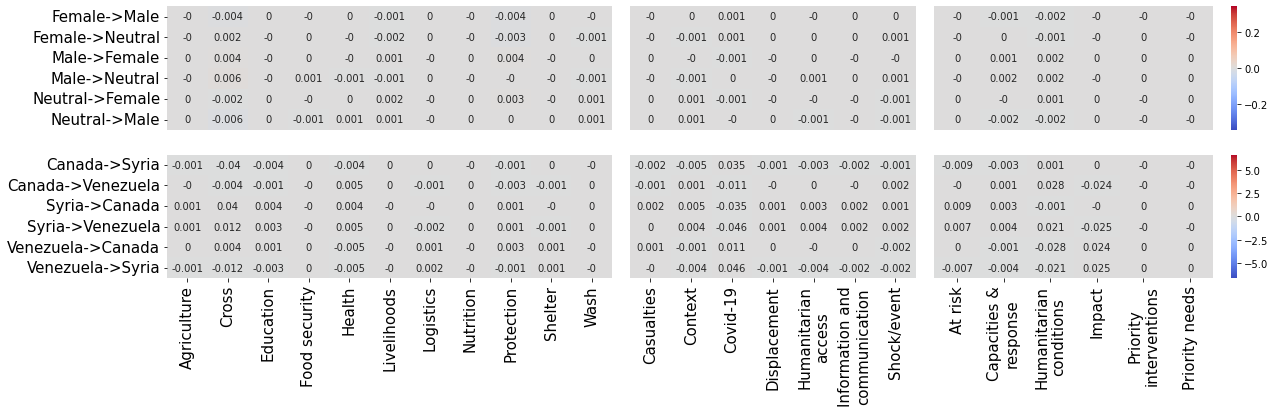}
  \end{subfigure}
  \caption{
  The results of $\text{Tag-Shift}$ bias metric for \modelbaseline architecture using the \modelhumbert as the backbone. (Top) Original model without debiasing; (Bottom) Counterfactual debiasing. 
  }
  \label{fig:results-humbert-base}
  \vspace{-4mm}
\end{figure*}

\begin{figure*}[h]
  \centering
  \begin{subfigure}[b]{1.0\textwidth}
    \centering
    \includegraphics[width=1\textwidth]{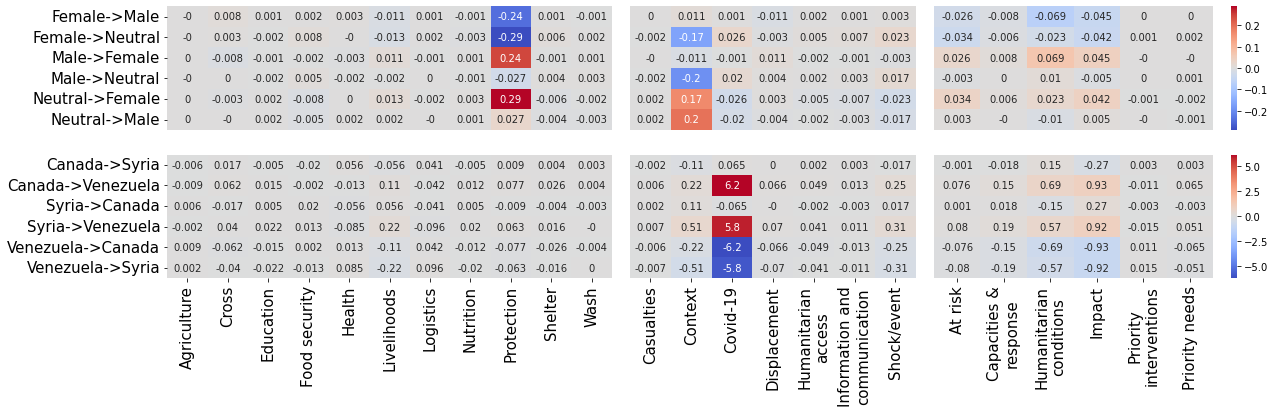}
  \end{subfigure}
  
  \begin{subfigure}[b]{1.0\textwidth}
    \centering
    \includegraphics[width=1\textwidth]{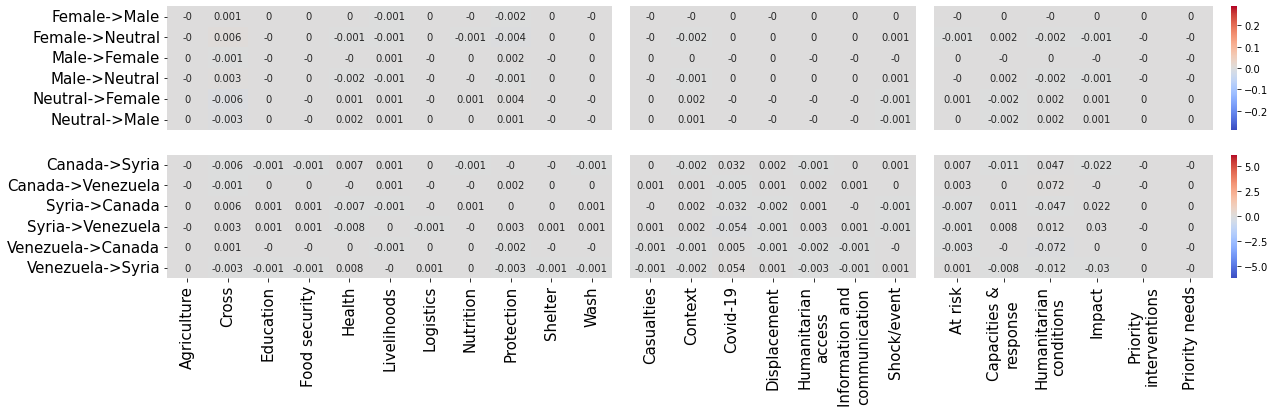}
  \end{subfigure}
  \caption{
  The results of $\text{Tag-Shift}$ bias metric for \modelours architecture using the \modelmbert as the backbone. (Top) Original model without debiasing; (Bottom) Counterfactual debiasing. 
  }
   \label{fig:mbert-ours}
  \vspace{-4mm}
\end{figure*}

Figures~\ref{fig:results-humbert-base}-\ref{fig:mbert-ours} report the results of gender and country bias measurement over various backbone LLMs, and architectures, before after applying CDA bias mitigation.

\begin{figure*}[h]
  \centering
  \begin{subfigure}[b]{1.0\textwidth}
    \centering
    \includegraphics[width=1\textwidth]{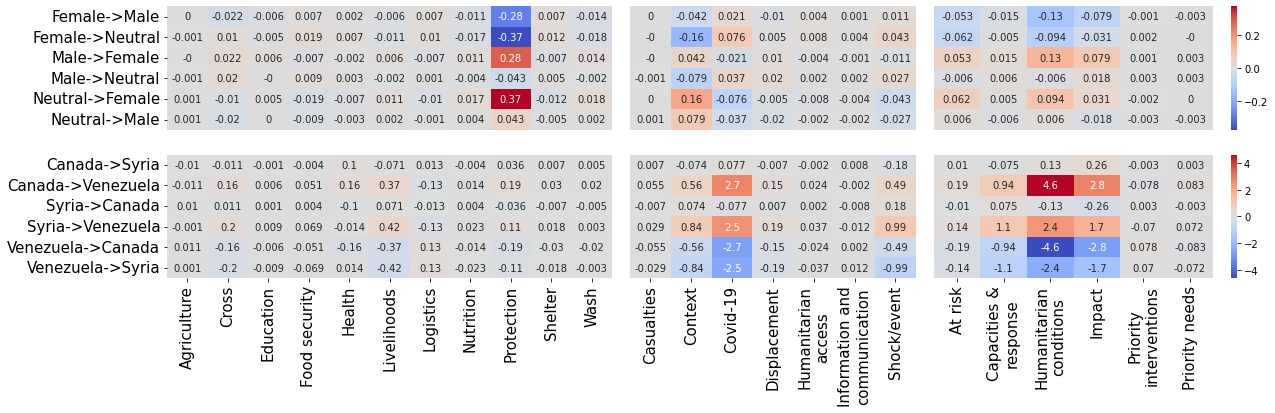}
  \end{subfigure}
  
  \begin{subfigure}[b]{1.0\textwidth}
    \centering
    \includegraphics[width=1\textwidth]{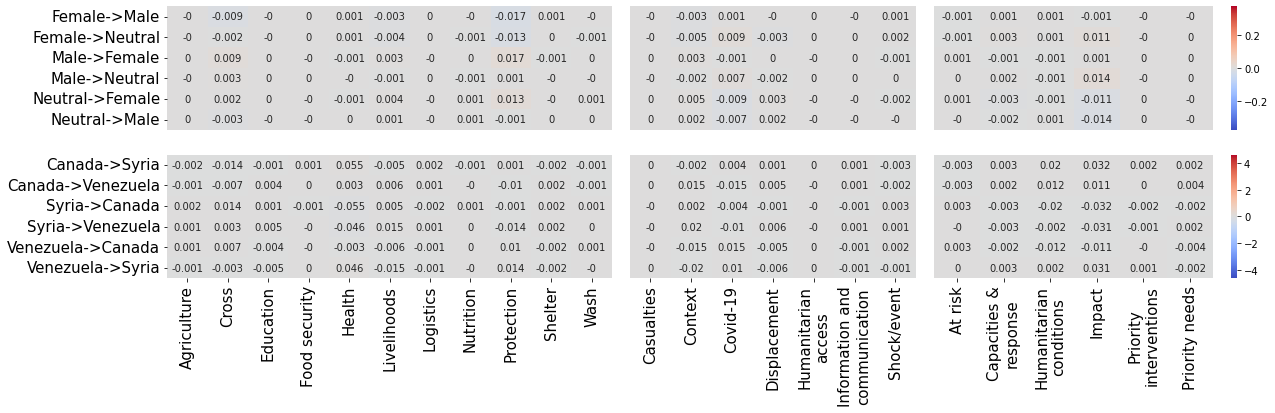}
  \end{subfigure}
  \caption{
  The results of $\text{Tag-Shift}$ bias metric for \modelbaseline architecture using the \modelxlm as the backbone. (Top) Original model without debiasing; (Bottom) Counterfactual debiasing. 
  }
  \vspace{-4mm}
\end{figure*}


\begin{figure*}[h]
  \centering
  \begin{subfigure}[b]{1.0\textwidth}
    \centering
    \includegraphics[width=1\textwidth]{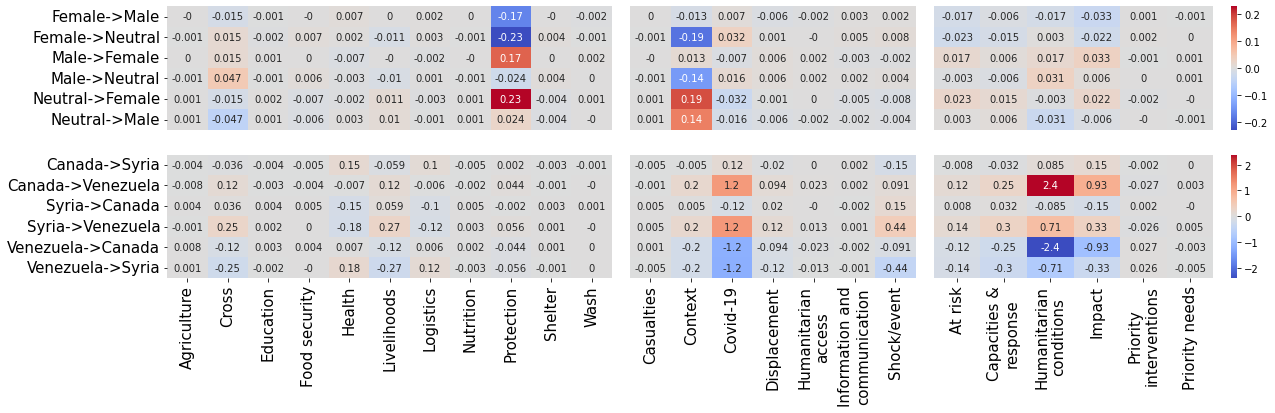}
  \end{subfigure}
  
  \begin{subfigure}[b]{1.0\textwidth}
    \centering
    \includegraphics[width=1\textwidth]{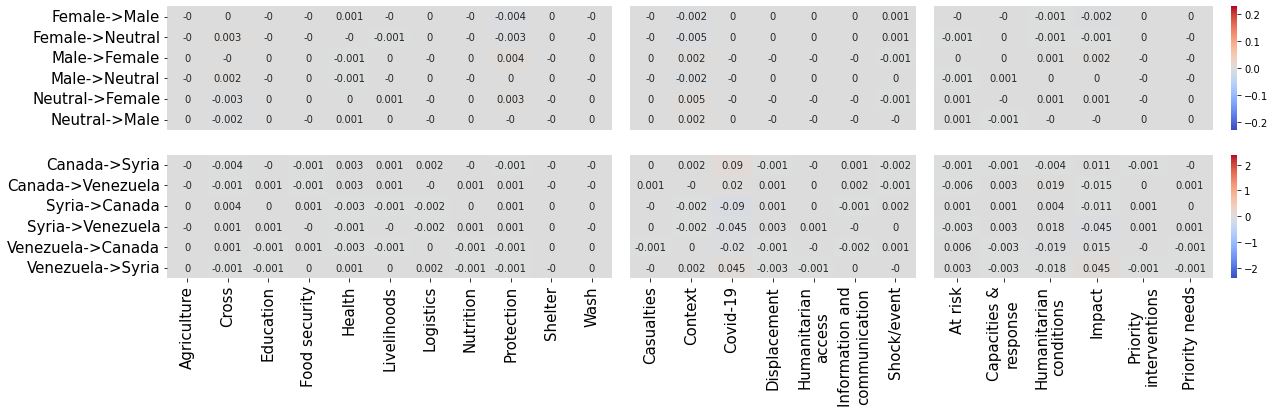}
  \end{subfigure}
  \caption{
  The results of $\text{Tag-Shift}$ bias metric for \modelours architecture using the \modelxlm as the backbone. (Top) Original model without debiasing; (Bottom) Counterfactual debiasing. 
  }
  \vspace{-4mm}
\end{figure*}



\begin{figure*}[h]
  \centering
  \begin{subfigure}[b]{1.0\textwidth}
    \centering
    \includegraphics[width=1\textwidth]{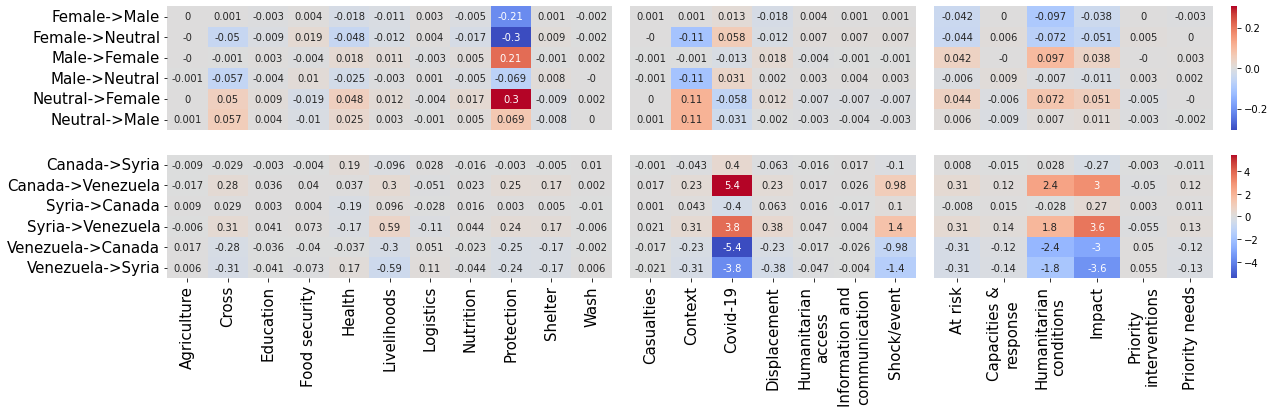}
  \end{subfigure}
  
  \begin{subfigure}[b]{1.0\textwidth}
    \centering
    \includegraphics[width=1\textwidth]{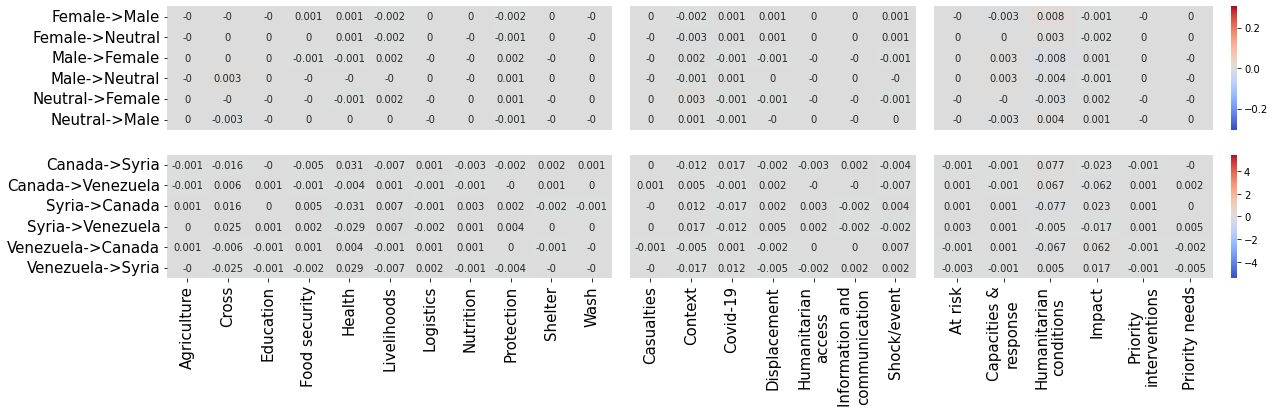}
  \end{subfigure}
  \caption{
  The results of $\text{Tag-Shift}$ bias metric for \modelbaseline architecture using the \modelmbert as the backbone. (Top) Original model without debiasing; (Bottom) Counterfactual debiasing. 
  }
  \vspace{-4mm}
\end{figure*}


\end{document}